\documentclass[10pt, conference, compsocconf]{IEEEtran}
\usepackage{url}
\usepackage{caption}
\usepackage{array}
\usepackage{textcomp}
\usepackage{amsfonts,amsmath,amssymb}
\usepackage{bbm}
\usepackage{multirow}
\usepackage{subfigure}
\usepackage{graphicx}
\usepackage{algorithmic,algorithm}
\usepackage{bm}

\newtheorem{defi}{Definition}
\DeclareMathOperator{\Tr}{tr}
\DeclareMathOperator{\card}{card}
\begin{document}
\setlength{\textfloatsep}{10pt plus 1.0pt minus 4.0pt}
\title{Multilabel Consensus Classification}
\author{Sihong Xie$^\dag$
\and 
Xiangnan Kong$^\dag$
\and 
Jing Gao$^\S$
\and 
Wei Fan$^\ddag$
\and
Philip S. Yu$^\dag$}
\maketitle
\begin{abstract}
\let\thefootnote\relax\footnotetext{
$^\dag$Department of Computer Science, University of Illinois at Chicago\hspace{.1in}
$^\S$Department of Computer Science and Engineering, University at Buffalo\hspace{.1in}
$^\ddag$Huawei Noah's Ark Lab, Hong Kong\hspace{.1in}}
In the era of big data, a large amount of noisy and incomplete data can be collected from multiple sources for prediction tasks.
Combining multiple models or data sources helps to counteract
the effects of low data quality and the bias of any single model or data source,
and thus can improve
the robustness and the performance of predictive models.
Out of privacy, storage and bandwidth considerations,
in certain circumstances
one has to combine the predictions from multiple models
or data sources to obtain the final predictions without accessing the raw data.
Consensus-based prediction combination algorithms are effective for such situations.
However, current research on prediction combination focuses on the single label setting,
where an instance can have one and only one label.
Nonetheless, data nowadays are usually multilabeled,
such that more than one label have to be predicted at the same time.
Direct applications of existing prediction combination methods
to multilabel settings can lead to degenerated performance.
In this paper, we address the challenges of combining predictions from multiple multilabel classifiers
and propose two novel algorithms, MLCM-r (\underline{M}ulti\underline{L}abel \underline{C}onsensus \underline{M}aximization for ranking)
and MLCM-a (MLCM for microAUC).
These algorithms can capture
label correlations that are common in multilabel classifications,
and optimize corresponding performance metrics.
Experimental results on popular multilabel classification tasks verify
the theoretical analysis and effectiveness of the proposed methods.
\end{abstract}

\section{Introduction}
Combining multiple models or data sources has been attracting more and more attentions
in data mining and machine learning research communities.
Real-world data are usually massive, noisy and incomplete.
To improve the robustness and generalization ability of learning methods on these real-world data,
one has to combine multiple models and exploit the knowledge of multiple data sources.
Many methods have been proposed for the purpose, such as~\cite{Schapire2002, Breiman1996},
which focus on learning ensembles of models from the training data and predictions on test data.
Due to privacy, bandwidth or storage issues, there are situations
where
we cannot have access to either the training data nor the testing data directly.
Instead, only the predictions of base models are available.
For example, in finance, aggregating customers information from multiple banks would benefit customer segmentation analysis.
However, it would be unsafe or infeasible to transfer the customer information across different banks.
One solution to this problem is that
we can apply the analysis at each bank individually, and then aggregate the predictions from multiple banks.
Prediction combination is a powerful paradigm for such situations
with an abundance of studies~\cite{jing09, li08consensus, li07, strehl03, wang09}.
These algorithms
combine the predictions of multiple supervised and/or unsupervised models,
in hope of improving accuracies by exploiting the strengths of different models or data sources,
without access to training or test data.

Conventional research on prediction combination
has been focusing on single label classification
and cannot handle multilabel classification.
Meanwhile, multilabel classification has seen its wide application in text/image categorization,
bioinformatics and so on, and therefore is of practical importance.
Although certain ensemble methods~\cite{Yu2012, Shi2011, Zhang2010} have been proposed to handle multilabel classification,
they focus on building the ensemble from training data, not on prediction combination.
Given the practical needs to combine multilabel predictions from multiple models/data sources without training and test data,
we identify the following challenges that need to be addressed in order to bridge the gap.
First,
although
state-of-the-art multilabel classification methods show that label correlations
can help improving classification performances,
how to exploit label correlations solely using predictions of base models
has not been addressed before.
Second,
there are various evaluation metrics for multilabel classification,
such as microAUC, ranking loss, one error, etc.~\cite{Dembczynski12,Elisseeff01},
it is more desirable to design algorithms that can be proved to be optimal for a specific metric,
as different applications require different quality measures.
Although, in~\cite{Dembczynski12},
they pointed out that optimizing different metrics
translates into the modeling of different label correlations,
it is non-trivial to align prediction combination methods with the modeling of label correlation
in order to optimize a specific metric.
There is no existing work that addresses the above issues.

In this paper, we address the above challenges by proposing two different algorithms that can model
label correlations given only the predictions of base models.
The algorithms are designed and proved to optimize two widely used but fundamentally different evaluation metrics, respectively.
The first algorithm MLCM-r
consolidates the predictions of base models via 
maximizing model consensus and
exploits label correlations using
random walk in the label space.
The algorithm is proved
to optimize ranking loss,
which measures the quality of the predictions on a per instance basis (e.g. find relevant labels for a query in image search engine).
Another important multilabel performance metric is microAUC,
which treats all instances combined as a single prediction task (e.g. find tags to describe a set of images.
Section~\ref{sec:mlcma} describes how microAUC differs from ranking loss).
Since a model that optimizes ranking loss might not be optimal on microAUC,
it is necessary to develop an alternative model that can combine predictions to optimize microAUC.
We propose a second algorithm called
MLCM-a (MultiLabel Consensus Maximization for microAUC)
for this purpose.
MLCM-a is formulated as a optimization problem that regularizes prediction consolidation using partial correlations between labels,
and we show that the objective of this formulation optimizes microAUC.

The contributions of this paper can be summarized as follows.
\begin{itemize}
\item We first study the problem of how to
combine predictions of multiple models in multilabel learning
without access to training and test data.
\item 
We propose two novel algorithms that can jointly model correlations among different labels
and the consensus among multiple models.
As different applications require different multilabel performance metrics,
we prove that the two algorithms optimize two multilabel classification metrics, respectively.
As far as we know, this is the first work that addresses the
multilabel-label consensus learning problem
and optimize particular metrics.
\item We compare the proposed models to 3 baselines on 6 multilabel classification tasks,
with a maximum of 45\% percent of reduction in ranking loss and 20\% percent of increase in microAUC.
\end{itemize}

\section{Preliminary}
In this section we recapitulate model combination
and multilabel classification algorithms, along with the challenges that we are addressing.
Table~\ref{tab:notations} summarizes most of the symbols and their definitions used in this paper.
We use boldface lower-case letters for vectors (e.g., $\mathbf{x}$) and capital letters for matrices (e.g., $Y$).
\begin{table}
\caption{Notations}
\label{tab:notations}
\begin{tabular}{c|c}
\hline
Symbol & Meaning\\\hline
$m$ & Number of multilabel classifiers\\
$n$ & Number of instances\\
$l$ & Number of labels\\
$\mathbf{x}$ & An instance\\
$\mathbf{z}$ & Ground truth labels of $\mathbf{x}$\\
$Y^{k}$ & output of the $k$th model\\
$\bar{Y}$ & Average of $Y^{k},k=1,\dots,m$\\
$\mathbf{y}^{k}_{i}$ & prediction of the $k$th model for the $i$th instance\\
$Y$ & Consolidated prediction of $Y^{k},k=1,\dots,m$\\
$\left<\cdot,\cdot\right>$ & inner product of two vectors\\
$|\cdot|$ & Determinant of a matrix\\
$\|\cdot\|$ & Frobenius norm of a matrix\\
$\mathbbm{1}[\cdot]$ & indicator of a predicate\\
$\card(A)$ & cardinality of the set $A$\\
\hline
\end{tabular}
\end{table}

\subsection{Multilabel Classification}
In multilabel-label classification problems, the data are in the form of $(\mathbf{x}, \mathbf{z})$,
where $\mathbf{x}$ is the feature vector of an instance and $\mathbf{z}$ is the label vector.
Suppose $L$ is the set of all $\ell$ possible labels, then $\mathbf{z}$ is a vector with length $|L|=l$ and
$z_{\ell}\in\{0,1\}$ denotes the value of the $\ell$-th label.
Multilabel classification is different from multiclass classification.
In multiclass classification, an instance have only one label, which can take more than two values (or classes).
However, in multilabel classification,
an instance can have more than one label,
each of which can take one and only one of the multiple values (classes).
For example, an account on a social network (LinkedIn, Facebook, etc.) can have multiple labels
such as ``sex'' and ``is employed'',
while there can be only one specific value for the label ``is employed''.
Multilabel classification introduces various unique challenges,
such as sparsity and imbalance of labels, multiple performance metrics of a model, etc.
Among these challenges, how to model and exploit label relationships to improve accuracy
has been studied intensively in~\cite{Dembczynski10,Petterson10,Read09,TsoumakasKV11,Zhang10}.
There are various types of label relationships,
the simplest one is pair-wise correlation, which specifies
how often two labels co-occur.
There are also some more complicated label relationships, such as hierarchical organizations of labels or high order relationships.
Recently, certain types of label relationship is shown to be connected to certain corresponding evaluation metrics.
For example, it is shown in~\cite{Dembczynski12} that
if one can compute the relevance score of each individual label given an instance,
the ranking according to the scores would yield the minimum ranking loss.
Conventional multilabel classification algorithms mainly focus on how to exploit label correlations
from training data.
These methods cannot directly address the challenge of combining multilabel predictions
without access to training or test data.

\subsection{Prediction Combination Algorithms}
Given the predictions of multiple models,
one needs to combine the predictions in order to obtain
a single final prediction.
Suppose there are $m$ base models,
whose predictions can be denoted by $\{Y^{1},\dots, Y^{m}\}$.
For $k=1,\dots, m$,
$Y^{k}$ is an $n\times l$ matrix, the $(i,\ell)$ element $Y_{i\ell}$ gives the class value
of the $i$-th instance for the $\ell$-th label, according to the $k$-th model.
$Y^{k}$ is a binary matrix
specifying the presence of a label in an instance.
The simplest form of model combination is majority voting,
where each base model votes in favor of a certain class for each label.
Afterwards,
instances are classified based on the votes an instance receives for different classes.
Formally, majority voting estimates the posterior probability of seeing a label given an instance, $p(y_{i\ell}|\mathbf{x}_{i})$
using the average of $Y_{i\ell}^{k}, k=1,\dots, m$.
Without access to the training and test data, the final predictions can be improved
by exploiting the correlations of predictions between two instances,
as it has been done in~\cite{jing09, li08consensus, li07, strehl03}.
However, under multilabel settings,
the correlation between different labels is also an important piece of information to exploit.
To achieve the best prediction combination results,
one should properly consider and model all available information, namely,
the correlations between labels and those between predictions of different instances.
The lack of existing model calls for novel models for this non-trivial task.
Furthermore, there are many multilabel classification performance metrics to choose from,
depending the application at hand.
Different metrics require fundamentally different ways of modeling of
label correlation~\cite{Dembczynski12},
and how to align different label correlations with various existing prediction combination methods
to optimize a certain performance metric
is another challenge that has not been explored before.

\section{Problem Formulation}
Given the multilabel predictions of base classifiers that do not necessarily consider label correlations
during training or testing,
we wish to produce improved consolidated predictions via
explicitly modeling label correlations without access to the training or test data.
Since the improvements in multilabel performance can be measured in many different metrics,
we further wish we can choose the right algorithm from a set of algorithms to optimize the desired metric.
We propose a family of algorithms as a solution to the above problem.
The algorithms can infer label correlations and combine predictions simultaneously,
and more importantly, optimize two different multilabel performance metrics,
namely,
ranking loss (Section~\ref{sec:mlcmr}) and
microAUC (Section~\ref{sec:mlcma}).

%
\section{Multilabel Consensus Maximization for Ranking Loss}
\label{sec:mlcmr}

\subsection{Prediction combination based on model consensus}
\label{sec:bgcm}
\begin{figure}[t]
\centering
\includegraphics[width=2.7in]{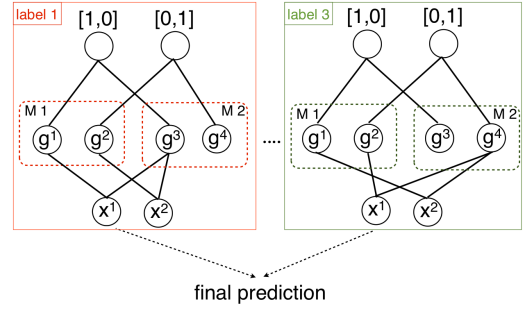}
\caption{Applying BGCM to multilabel prediction combination}
\label{fig:bipartite_graph_1}
\end{figure}

Prediction combination methods
have been explored in many previous works~\cite{jing09, li07, li08consensus, strehl03, wang09}.
Nonetheless, these methods are designed for single-label, multiclass classification,
such that prediction combination happens within individual labels.
A trivial way to apply these methods to multilabel tasks is to first
combine the predictions of base models for each label,
resulting in the consolidated predictions for {\it individual} labels.
Then these preliminary consolidated predictions
are pooled together as the final prediction for the multilabel task.
This process
treats labels independently without exploiting label correlations,
and is similar to the Binary Relevance (BR) method in multilabel classification literature.
To illustrate these concepts,
we take one of the multiclass prediction combination algorithms, BGCM~\cite{jing09}, as an example.
BGCM seeks consolidated predictions that are agreed upon by the base models as much as possible.
Without loss of generality, we assume that the base models are supervised models.
In particular, 
for {\it each} label,
BGCM constructs a bipartite graph to represent the predictions of base classifiers.
An example of applying BGCM to a prediction combination task with 2 instances, 2 classifiers, 2 classes of each of the 3 labels
is shown in
Figure~\ref{fig:bipartite_graph_1} (only the schemas of the first and last labels are shown).

In general,
given the predictions of $m$ classifiers for $n$ instances, with $c$ classes from a single label,
the bipartite graph has $n$ instances nodes and $v=m\times c$ group nodes.
In the above example, the bipartite graph for the first label
is shown
in the left rectangle,
where group nodes are annotated with the letter $g$ and instance nodes
with the letter $x$.
Each node is associated with a probability distribution over $c$ classes (not shown in Figure~\ref{fig:bipartite_graph_1}).
The distribution for the $i$-th instance node is given by the row vector $\mathbf{u}_{i}, i=1,\dots, n$,
which are collectively denoted by the $n\times c$ matrix
$U=[\mathbf{u}_{1}^{\prime},\dots, \mathbf{u}_{n}^{\prime}]^{\prime}$.
Similarly, let the $v\times c$ matrix $Q=[\mathbf{q}_{1}^{\prime},\dots, \mathbf{q}_{v}^{\prime}]^{\prime}$
be the distributions of $v$ group nodes.
The connections of these nodes are determined by the predictions of the base models.
If $\mathbf{x}_i$ is classified into the $j$-th class by $k$-th model,
the $i$-th instance node is connected to the $(k-1)\times c+j$ group nodes.
In the above bipartite graph for label 1,
instance $\mathbf{x}_{1}$ is classified into class 1 by model 1,
then the first instance node is connected to $g^{1}$.
For a group node to represent a class,
it is connected to a class node.
Specifically, if a group node represents the $j$-th class, then it is connected to a class node
with class distribution $\mathbf{b}_j$, which has 1 at its $j$-th position and 0 otherwise.
Let the $v\times c$ matrix $B=[\mathbf{b}_{1}^{\prime},\dots,\mathbf{b}_{v}^{\prime}]^{\prime}$.
For example, since group nodes $g^{1}$ and $g^{3}$ respectively represent class 1 for two models,
they are connected to the first group node on the top row, with class distribution $[1,0]$.

For each label,
BGCM
solves the following
optimization problem 
to achieve maximal consensus among base models,
\begin{eqnarray}
\label{eq:cm_objective}
 \displaystyle\min_{U,Q}& \displaystyle\sum_{i=1}^{n}\sum_{j=1}^{v}a_{ij}\|\mathbf{u}_{i}-\mathbf{q}_{j}\|^{2}+\alpha\sum_{j=1}^{v}\|\mathbf{q}_{j}-\mathbf{b}_{j}\|^{2} \label{eq:consensus_loss}\\
 \mbox{s.t.} & u_{i\ell}\geq 0, \sum_{\ell=1}^{c}u_{i\ell}=1,i=1,\dots,n \label{eq:u_constraint}\\
 	     & q_{j\ell}\geq 0, \sum_{\ell=1}^{c}q_{j\ell}=1,j=1,\dots,v \nonumber
\end{eqnarray}
In Eq.(\ref{eq:cm_objective}), $a_{ij}=1$ indicates that the $i$-th instance node and the $j$-th group node
are connected, otherwise $a_{ij}=0$.
After the optimization problem is solved, the consolidated prediction of the $i$-th instance for a single label can be obtained by
taking the maximal value in $\mathbf{u}_{i}$.
The solution of the optimization problem achieves maximal consensus among base models,
an objective also pursued by other consensus based prediction combination methods~\cite{li07, li08consensus, strehl03}.
Although this objective can lead to the improvement of performance over base models,
these methods can only combine multilabel predictions by first combine the predictions for each label,
and then concatenate the predictions for individual labels to obtain the final prediction for multiple labels.
Apparently, no label correlation is modeled in this paradigm.
Next we propose a novel method based on BGCM to incorporate label correlations in multilabel predictions combination.

\subsection{MLCM-r}
\label{sec:mlcm_r}
\begin{figure}[t]
\subfigure[MLCM-r]{
		\includegraphics[width=1.5in]{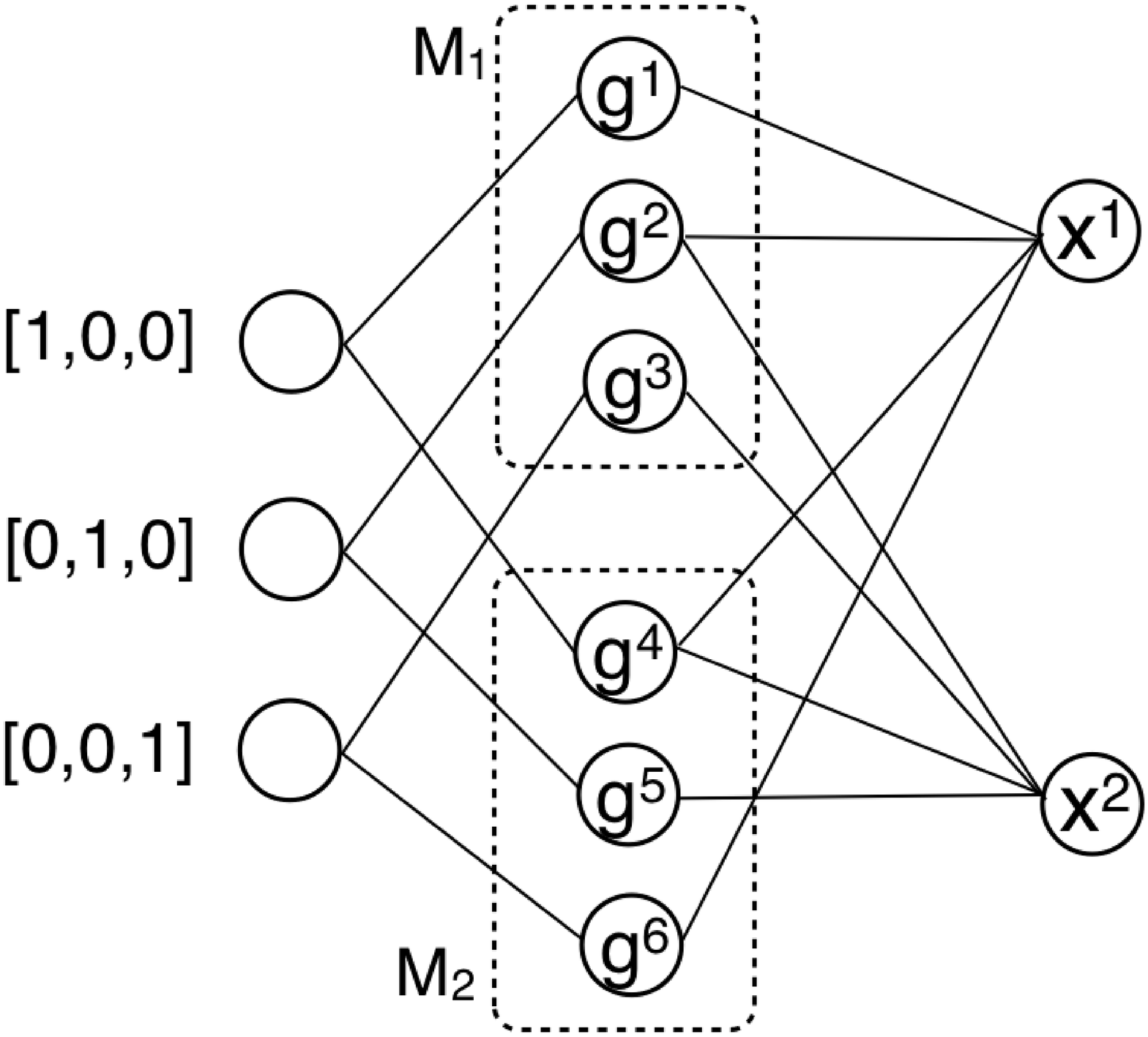}
                \label{fig:mlcmr}
}
\subfigure[Graph of group nodes encoding label relationships]{
		\includegraphics[width=1.5in]{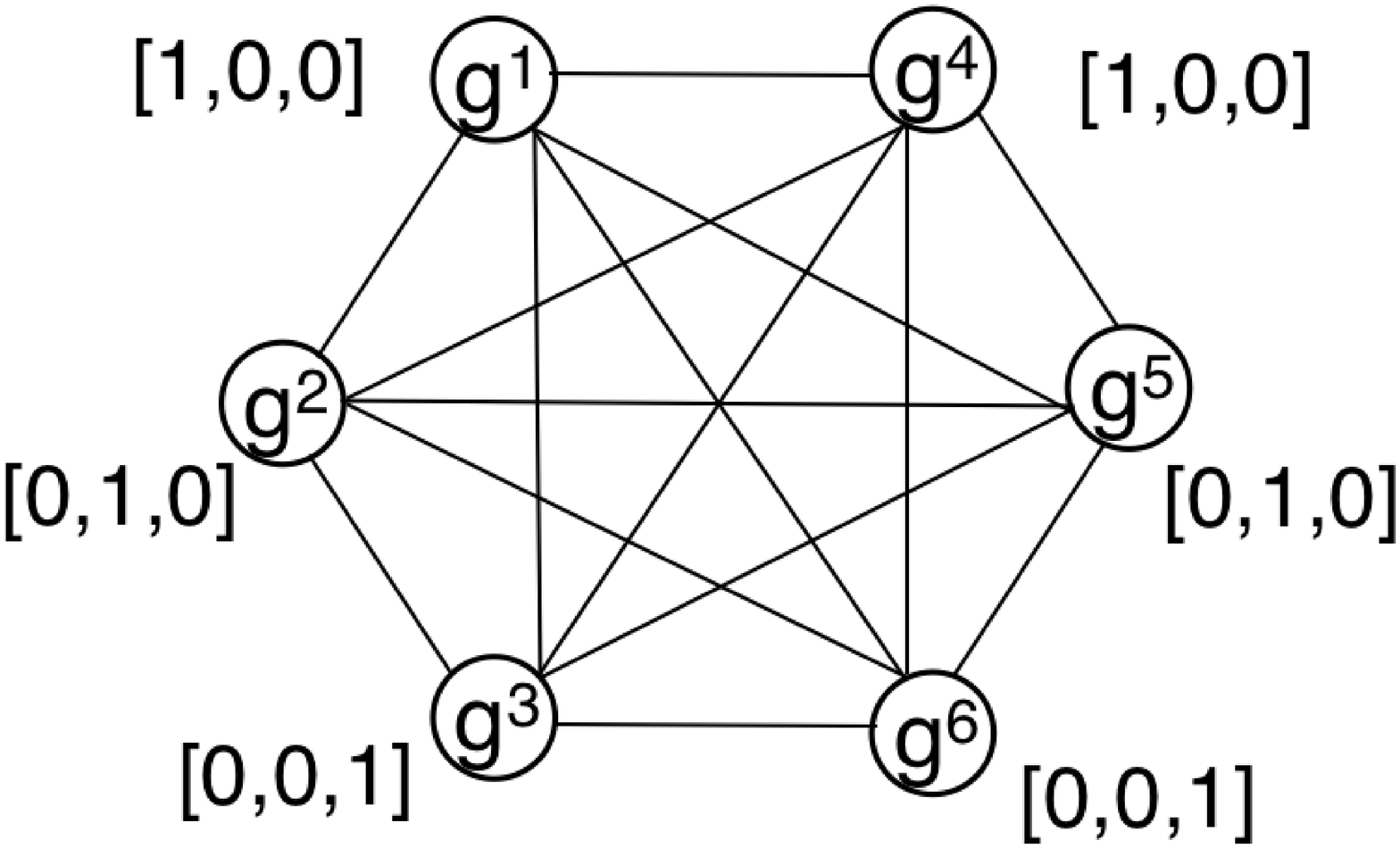}
                \label{fig:group_graph}
}
\caption{Bipartite graph for MLCM-r and its collapse to group nodes}
\end{figure}

According to the last section, one might wish to jointly model label correlations and model consensus to
overcome the drawback of BGCM under multilabel settings
while
exploiting the power of BGCM in maximizing model consensus.
We propose MLCM-r, which adopts the architecture of BGCM to achieve this goal.
For simplicity, we assume that each label consists of two classes.
We abuse the notations introduced in Section~\ref{sec:bgcm}.
In particular, we let the $n$ by $v$ $(v=m\times l)$ connection matrix $A$
encode the multilabel predictions,
where the $(i,(k-1)\times l+j)$-th entry is 1 if the $k$-th model predicts that
the $i$-th instance takes class 1 on the $j$-th label, otherwise the entry is 0.
Viewing $A$ as a connection matrix between instances and labels,
a bipartite graph can be constructed for MLCM-r.
An example of the bipartite graph of MLCM-r for 2 instances, 3 classes and two base multilabel classifiers
is shown in Figure~\ref{fig:mlcmr}.
Similar to the bipartite graph, the bipartite graph for MLCM-r has both group nodes and instance nodes,
annotated by the letters $g$ and $x$.
However, there are some differences between two bipartite graphs.
Surrounded by a rectangle with dashed line are
the group nodes from a classifier (e.g. the rectangle $M_{1}$ includes the group nodes from the first classifier).
A group node in Figure~\ref{fig:mlcmr} represents a label instead of a class in Figure~\ref{fig:bipartite_graph_1}.
An instance node in Figure~\ref{fig:mlcmr} can be connected to more than one group nodes from a classifier, naturally
representing the multilabel predictions.
These differences between Figure~\ref{fig:mlcmr} and Figure~\ref{fig:bipartite_graph_1}
bring more expressive power to MLCM-r, as summarized below:
\begin{itemize}
\item The connections between an instance and {\it all} labels are fully given by a single graph
in MLCM-r, instead of being broken down into multiple
bipartite graphs in BGCM.
\item most importantly, the relationship between labels can be derived in MLCM-r using Figure~\ref{fig:mlcmr},
as shown in the graph of group nodes in Figure~\ref{fig:group_graph}.
We give more details of this property of MLCM-r in Section~\ref{sec:mlcmr_analysis}.
\end{itemize}

According to the newly defined $A$, we
re-define the distributions associated with the nodes.
$u_{i\ell}$
(the $\ell$-th entry of $\mathbf{u}_{i}$)
is now defined to be the probability of the $i$-th instance taking class 1 on the $\ell$-th label.
Similarly $q_{j\ell}$ is defined as
the probability of seeing the $\ell$-th label given the $j$-th label (the reason of this definition is explained in the next section).
If the $j$-th group node represents the $\ell$-th label,
it is connected to a label node with distribution $\mathbf{b}_{j}$, which has
$1$ on its $\ell$-th entry and $0$ for the other entries,
let $B=[\mathbf{b}_{1}^{\prime},\dots,\mathbf{b}_{v}^{\prime}]^{\prime}$ similarly as in BGCM.
With the re-defined variables and constants (see Table~\ref{tab:notation_2}),
MLCM-r maximizes model consensus by solving
a similar optimization problem in Eq.(\ref{eq:consensus_loss}).
The closed form optimal solution is given in Eq.(\ref{eq:cm_closed_sol_q}) and Eq.(\ref{eq:cm_closed_sol_u}),
which
infer and exploit label correlations to minimize ranking loss, as analyzed in the next section.
\begin{table}[t]
\caption{Notations for MLCM-r}
\label{tab:notation_2}
\begin{tabular}{c|c}
\hline
Symbol & Meaning\\\hline
$A$ & $a_{i,j}$ is the prediction of label ($j$ mod $l$) on $\mathbf{x}_{i}$ by the $\lfloor j/l\rfloor$\\
$B$ & Label node class distribution\\
$U$ & $u_{i\ell}$ is the probability that label $\ell$ is relevant to $\mathbf{x}_{i}$\\
$Q$ & $q_{j\ell}$ is the probability of seeing label $\ell$ given label $j$\\
$I_{k}$ & $k$ dimensional identity matrix\\\hline
\end{tabular}
\end{table}
\subsection{Analysis of MLCM-r}
\label{sec:mlcmr_analysis}
In this section, we first analyze the property of MLCM-r, which is shown to perform a random walk in label space and
thus infer label correlations.
Then we introduce ranking loss, which is connected to MLCM-r to show that MLCM-r indeed optimizes ranking loss.

According to~\cite{jing09},
a closed form solution for the optimization problem is
\begin{equation}
\label{eq:cm_closed_sol_q}
Q^{\ast}=(I_{v}-D_{\lambda}D_{v}^{-1}A^{\prime}D_{n}^{-1}A)^{-1}D_{1-\lambda}B
\end{equation}
where $D_{v}=\text{diag}(\mathbbm{1}^{\prime}A)$, $D_{n}=\text{diag}(\mathbbm{1}^{\prime}A^{\prime})$,
$\mathbbm{1}$ is a column vector with all entries being 1.
$D_{\lambda}=D_{v}(\alpha I+D_{v})^{-1}$ and $D_{1-\lambda}=\alpha(\alpha I+D_{v})^{-1}$.
After $Q^{\ast}$ is obtained, $U^{\ast}$ is obtained using
\begin{equation}
\label{eq:cm_closed_sol_u}
U^{\ast}=D_{n}^{-1}AQ^{\ast}
\end{equation}
Eq.(\ref{eq:cm_closed_sol_q}) actually solves a problem similar
to personalized pagerank
over the graph in Figure~\ref{fig:group_graph}.
The graph consists of group nodes from Figure~\ref{fig:mlcmr}, with edges indicating strength
of connection between group nodes.
In particular, the graph expresses the chances of co-occurrence of two labels
in terms of the proportion of instances that have both labels simultaneously.
The results of the random walk is simply the probabilities that one node hits another node
during a specific random walk.
Since the nodes represent labels, the solution can be seen as the probabilities of seeing one label when starting from another label.
We analyze this intuition more formally.
We wish to establish the solution of the random walk $Q_{\ell j}^{\ast}$
as the probability of seeing the $j$-th label given the $\ell$-th label.
Fixing $j$ and looking at
Eq.(\ref{eq:cm_closed_sol_q})
in a column-wise perspective, for $j=1,\dots, v$,
we obtain
\begin{equation}
\label{eq:cm_closed_sol_q_colum}
Q^{\ast}_{\cdot j}=(I_{v}-D_{\lambda}D_{v}^{-1}A^{\prime}D_{n}^{-1}A)^{-1}D_{1-\lambda}B_{\cdot j}
\end{equation}
where $Q_{\cdot j}$ and $B_{\cdot j}$ are the $j$-th column of $Q$ and $B$,
$\lambda_{j}$ is the $j$-th diagonal entry of $D_{\lambda}$.
Let $S=D_{v}^{-1}A^{\prime}D_{n}^{-1}A$, which is the transition matrix.
Each row of $S$ is a probability distribution where $S_{ij}$ is the transition probability
from group node $i$ to group node $j$.
By the identity $(I-S)^{-1}=\sum_{t=0}^{\infty}S^{t}$,
we can re-write 
Eq.(\ref{eq:cm_closed_sol_q_colum}) as
\begin{equation}
\label{eq:cm_iter_sol_q}
Q^{\ast}_{\cdot j}=\left(\sum_{t=0}^{\infty}(D_{\lambda}S)^{t}\right)D_{1-\lambda}B_{\cdot j}
\end{equation}

Out of Eq.(\ref{eq:cm_iter_sol_q}),
we can construct a random walk where
a person takes from 0 to infinitely many steps to
eventually settle down at any one of the group nodes for label $j$
(note that there can be multiple group nodes for label $j$ given multiple classifiers,
e.g. group nodes $g^{1}$ and $g^{4}$ represent label $j$).
At each step, the person can choose to settle down with probability $1-\lambda_{i}$ at group node $i$,
or to take one more transition with probability $\lambda_{i}$,
given the current position being the $i$-th group node.
$(D_{\lambda}S)^{t}$ can be interpreted
similar to traditional random walk.
For the base case,
$(D_{\lambda}S)^{0}=I$ gives the probability that one starts from any one of the group nodes
and reaches {\it any} nodes in zero step.
Assume $((D_{\lambda}S)^{t-1})_{ij}$ is the probability that the person reaches node $j$ starting from node $i$ in $t-1$ steps.
Then $((D_{\lambda}S)^{t})_{ij}=\lambda_{i}\sum_{k=1}^{v}S_{ik}((D_{\lambda}S)^{t-1})_{kj}$,
which can be interpreted as the person chooses to continue walking with probability $\lambda_{i}$,
and ends up at node $j$ with probability $\sum_{k=1}^{v}S_{ik}((D_{\lambda}S)^{t-1})_{kj}$.
By induction, $(D_{\lambda}S)^{t}$ gives the probabilities of
moving from one node to another in $t$ steps without settling down.

Given the above interpretation and fixing $j=1$, we obtain
\begin{eqnarray}
((D_{\lambda}S)^{t}D_{1-\lambda}B)_{i1}&=&\sum_{k}((D_{\lambda}S)^{t})_{ik}(1-\lambda_{k})B_{k1}\nonumber\\
&=&\sum_{k}((D_{\lambda}S)^{t})_{ik}(1-\lambda_{k})\mathbbm{1}(B_{k1}=1)\nonumber
\end{eqnarray}
Note that $B$ is a matrix with 0 or 1 entries and $B_{k1}=1$ iff the $k$-th group node for label 1.
Also note that $(1-\lambda_{k})$ is the probability of settling down at the $k$-th group node.
Then a summand in the above summation is the probability of
starting from group node $i$
and settling down after $t$ steps of transition at the $k$-th group node belonging to label 1.
The sum of these probabilities is the probability that settling down at {\it any}
of the group nodes for label 1.
$Q^{\ast}_{\cdot 1}=(\sum_{t=0}^{\infty}(D_{\lambda}S)^{t})D_{1-\lambda}B_{\cdot 1}$,
and $Q^{\ast}_{\ell 1}$ gives the probability that, starting from the $\ell$-th group node, one reaches any group nodes of class 1.

According to Eq.(\ref{eq:cm_closed_sol_u}),
the $(i,\ell)$-th entry of $U$ is
\begin{eqnarray*}
& & U_{i\ell}=\frac{1}{d_{i}}\sum_{j=1}^{v}a_{ij}Q^{\ast}_{j\ell}\\
&=& \sum_{k=1}^{c}\frac{n_{k}}{d_{i}}\left(\sum_{j=1}^{v}\mathbbm{1}[B_{jk}=1]\frac{a_{ij}}{n_{k}}Q^{\ast}_{j\ell}\right)\\
&=&\sum_{k=1}^{c}p(k|\mathbf{x}_{i}) p(\ell|k,\mathbf{x}_{i})
\end{eqnarray*}
where $n_{k}=\sum_{j}a_{ij}\mathbbm{1}[B_{jk}=1]$, which is the total number
of group nodes of label $k$ that $\mathbf{x}_{i}$ connects to.
$p(k|\mathbf{x}_{i})=n_{k}/d_{i}$ is the probability that $\mathbf{x}_{i}$
has label $k$ according to $m$ base models.
$p(\ell|k,\mathbf{x}_{i})=\sum_{j=1}^{v}\mathbbm{1}[B_{jk}=1](a_{ij}/n_{k})Q^{\ast}_{j\ell}$
is simply the average of $Q^{\ast}_{j\ell}\mathbbm{1}[B_{jk}=1]$,
which is probability of going from label $k$ to label $\ell$.
These two probabilities depend on $\mathbf{x}_{i}$ due to the term
$d_{i}$ and $n_{k}$, which depend on the connectivity between $\mathbf{x}_{i}$ and the group nodes.
Therefore, MLCM-r computes the probabilities $p(y_{\ell}=1|\mathbf{x}_{i})$.

The above results connects MLCM-r to ranking loss, which is defined below.
Ranking loss measures how much the ranking of the labels violates the relevant-irrelevant relationship
between pairs of labels.
Let $P_{i}$ be the set of relevant labels for $\mathbf{x}_{i}$,
and $N_{i}$ the set of irrelevant labels.
$P_{i}\times N_{i}$ is the set of all pairs of relevant and irrelevant labels.
Given the relevance scores $f(\ell, \mathbf{x}_{i})$ of label $\ell$ of
$\mathbf{x}_{i}, \ell=1,\dots, c, i=1,\dots, n$, ranking loss is defined as
\begin{equation}
\text{ranking loss}=\sum_{i=1}^{n} \sum_{\ell\in P_{i},\ell^{\prime}\in N_{i}}\frac{\mathbbm{1}[f(\ell,\mathbf{x}_{i})\leq f(\ell^{\prime},\mathbf{x}_{i})]}{\card(P_{i}\times N_{i})}
\end{equation}
In~\cite{Dembczynski12}, it was proved that the expected ranking loss
is minimized by the ranks of the relevance scores, which is defined as the posterior probability
$p(y_{i\ell}=1|\mathbf{x}_{i})$.
In other words, so long as the probability $p(y_{i\ell}=1|\mathbf{x}_{i})$ can be estimated accurately,
one should be able to achieve a low ranking loss.
But this is what exactly MLCM-r does, as we show above.
Therefore we conclude that MLCM-r minimizes ranking loss.


\section{Multilabel Consensus Maximization for microAUC}
\label{sec:mlcma}
In this section, we propose another algorithm for multilabel prediction combination.
This algorithm differs from the first one in that it optimizes microAUC, which is
both theoretically and practically different from ranking loss.
After briefly review the differences between two metrics, we describe the second algorithm
based on simple averaging.

\subsection{microAUC and its properties}
\label{sec:micro_auc}
AUC (Area Under the Curve) is a binary classification metric for situations where one class greatly out-numbers the other class.
In multilabel classification, an instance usually has only a quite small proportion of all labels.
For example, in text tagging, there can be thousands of tags, yet an article usually has only a couple of tags.
Since there can be much more irrelevant label than relevant labels,
AUC can be adopted to the multilabel setting, where the metric is called microAUC.
Formally, the label matrix $Z=[\mathbf{z}_{1}^{\prime},\dots, \mathbf{z}_{n}^{\prime}]^{\prime}$ for $n$ instances
has a total of $n\times l$ entries.
Let $P$ be the set of positive (relevant) entries and $N$ the set of negative (irrelevant) entries, $\card(P)\ll \card(N)$.
Given a list of relevance scores of all entries,
microAUC~\cite{Corinna03,Hanley1982} is defined as
\begin{equation}
\text{microAUC}=\sum_{i\in P}\sum_{j\in N}\frac{\mathbbm{1}[f(i)>f(j)]}{\card(P)\times \card(N)}
\end{equation}
where $f(i)$ is the relevance score of entry $i$.
Observe that microAUC is the ratio between the number of correctly ordered pairs
and the total pairs.
A fundamental difference between two metrics is that,
ranking loss does not compare the ranks between labels of two different instances,
while microAUC compares the ranks of all possible pairs of labels, no matter they are from the same instance or not.
In this sense, approaches that optimize ranking loss does not necessarily optimize microAUC.
Next we introduce simple averaging,
based on which we propose an algorithm that combines multilabel predictions to directly optimize microAUC.

\subsection{Simple averaging}
Perhaps the simplest way to consolidate predictions from multiple models is to take
the average of the predictions:
\begin{equation}
Y={\bar Y} = \frac{1}{m}\sum_{k=1}^{m}Y^{k}=\sum_{k=1}^{m}Y^{k}(mI_{l})^{-1}
\label{eq:averaging}
\end{equation}
where $I_{l}$ is the $l$ dimensional identity matrix.
The loss function
that simple averaging minimizes
is the sum of squared error between
the consolidated prediction $Y$ and the base models' predictions $\{Y^{1},\dots,Y^{m}\}$.
Formally, we adopt the results
from~\cite{li08consensus}.
\begin{eqnarray}
&&\sum_{k=1}^{m}\|Y^{k}-Y\|^{2}=\sum_{i=1}^{n}\sum_{k=1}^{m}\|\mathbf{y}^{k}_{i}-\mathbf{y}_{i}\|^{2}\nonumber\\
&=&\sum_{i=1}^{n}\sum_{k=1}^{m}\|\mathbf{y}^{k}_{i}-\mathbf{\bar y}_{i}+\mathbf{\bar y}_{i}-\mathbf{y}_{i}\|^{2}\nonumber\\
&=&\sum_{i=1}^{n}\sum_{k=1}^{m}\|\mathbf{y}^{k}_{i}-\mathbf{\bar y}_{i}\|^2+\sum_{i=1}^{n}\|{\mathbf{\bar y}}_{i}-\mathbf{y}_{i}\|^{2}\label{eq:consensus_0}
\end{eqnarray}
The last equality follows from $\sum_{k=1}^{m}(\mathbf{y}^{k}_{i}-\mathbf{\bar y}_{i})=0$,
as $m\mathbf{\bar y}_{i}=\sum_{k=1}^{m}\mathbf{y}^{k}_{i}$.
Note that the first term in the last line has nothing to do with $Y$.
Therefore, the minimum of the sum of squared errors
is attained by taking $Y={\bar Y}$.

There have been several applications of simple averaging
in multilabel classification to combine the results of multiple models,
such as ECC~\cite{Read09}, Rakel~\cite{TsoumakasKV11}, Model-shared subspace boosting~\cite{Yan2007}
and BoosTexter~\cite{Schapire2002}.
In these methods, label dependencies are modeled in the training phase and the combination step
does not consider any label dependency.
Therefore, if base models fail to model label dependencies in training and testing phases,
simple averaging cannot reconstruct the label dependency information solely from the predictions.

\subsection{MLCM-a}
\label{sec:mlcm_a}
\begin{figure}
\subfigure[Ranking loss]{
		\includegraphics[width=1.65in]{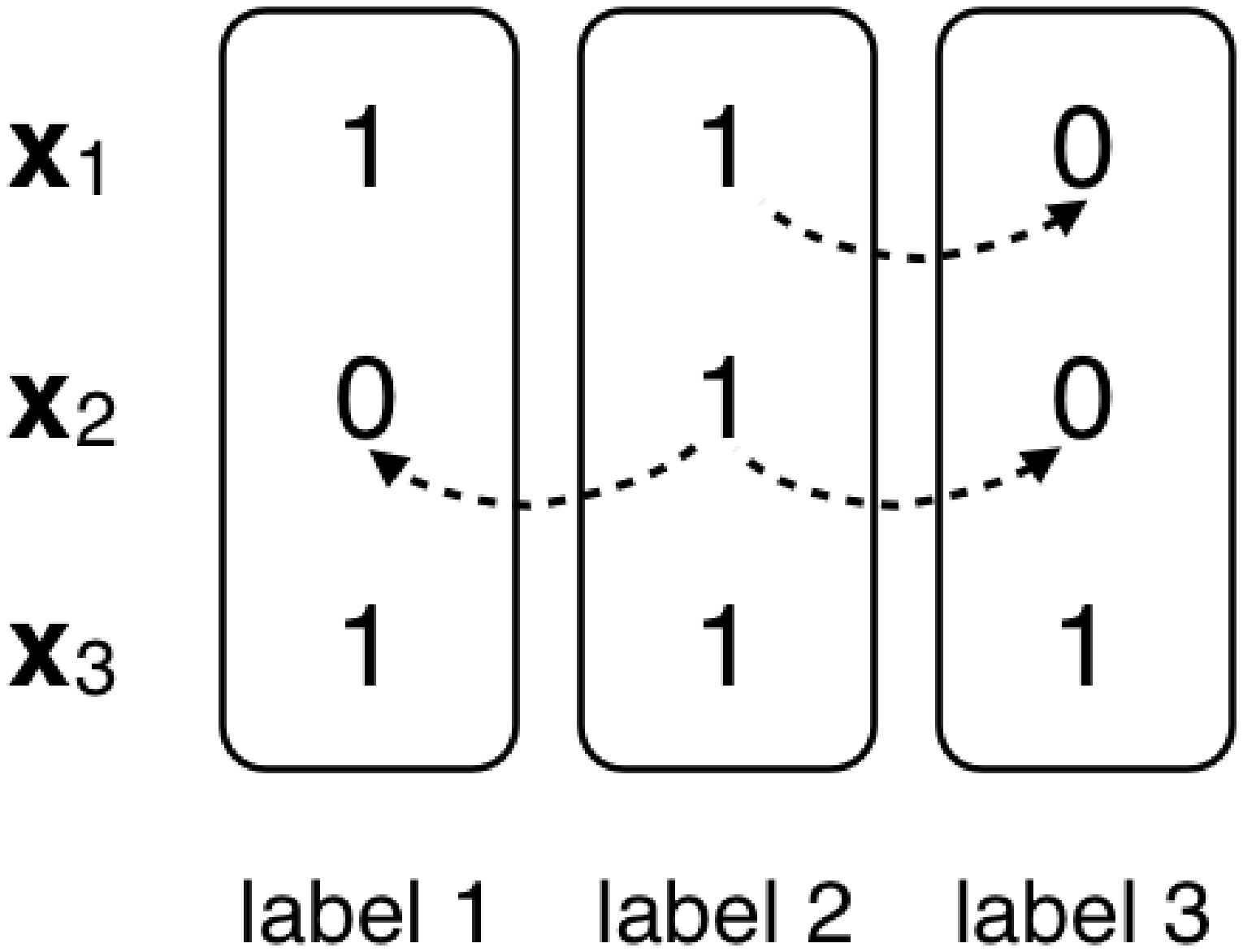}
                \label{fig:ranking_loss}
  }%
\subfigure[microAUC]{
		\includegraphics[width=1.6in]{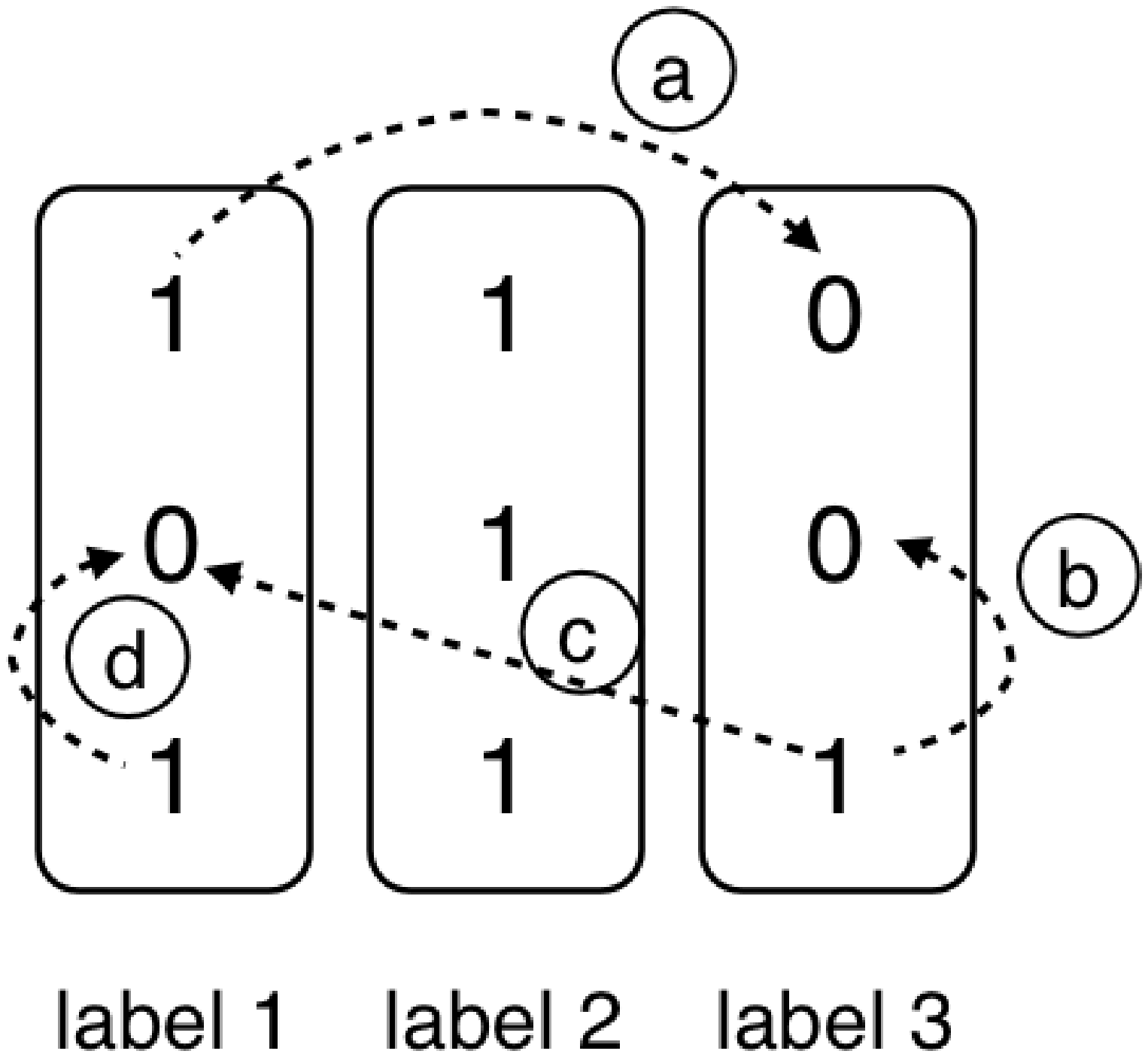}
                \label{fig:microAUC}
  }%
\caption{Comparison of ranking loss and microAUC}
\label{fig:comp_rank_auc}
\end{figure}

We examine microAUC more closely to motivate the method to be proposed.
In Figure~\ref{fig:comp_rank_auc}, we graphically demonstrate the differences
between ranking loss and microAUC.
Assume we have 3 labels and 3 instances $\{\mathbf{x}_{1},\dots,\mathbf{x}_{3}\}$.
The ground truth labels of the 3 instances are layed out as in the $3\times 3$ label matrix
$Z=[\mathbf{z}_{1}^{\prime},\dots, \mathbf{z}_{3}^{\prime}]^{\prime}$
where $\mathbf{z}_{i}$ is a row vector of the values of all labels for $\mathbf{x}_{i}$.
The values of the entries for a label are grouped in rectangles, while
each row represents the labels of an instance.
Ranking loss accounts the pairwise relationship between the labels {\it within} an instance.
Therefore, in Figure~\ref{fig:ranking_loss}, 3 pairs of relative ranks of entries will contribute to the ranking loss,
as indicated by arrows pointing from relevant labels to irrelevant ones within each instance.
However, there are more pairs of entries that microAUC accounts for.
Given a relevant label for an instance, microAUC pairs it with {\it all} other irrelevant labels of all instances, including itself.
In Figure~\ref{fig:microAUC},
example pairs of relevant and irrelevant entries are indicated by arrows labeled by letters.
We do not draw all pairs in Figure~\ref{fig:microAUC} to avoid untidiness.
Note that arrow {\bf a} indicates the sort of pairs of entries considered by ranking loss.
Arrow {\bf b} indicates pairs of entries within a label for different instances,
and arrow {\bf c} points from a label of an instance to a different label of a different instance.

Pairs of entries indicated by arrow {\bf b} or {\bf d} must have been handled by any reasonable base models,
which predict the relevance of a label to the instances.
Pairs of entries indicated by arrow {\bf a} consist only a small portion of all pairs if $n$ is large,
due to the sparsity of relevant labels.
Therefore, the major challenge in optimizing microAUC is
how to enforce preference of one label over other labels across different instances, such as the pairs indicated by arrow {\bf c}.
Without loss of generality, taken Figure~\ref{fig:microAUC} as an example,
given two instances $\mathbf{x}_{2}$ and $\mathbf{x}_{3}$,
we need to estimate the posteriors $p(y_{j}=1|\mathbf{x}_{i})$ for $i\in\{2,3\}, j\in\{1,3\}$,
in order to derive preferences between relevant and irrelevant labels.
Suppose with high probability that $p(y_{1}=1|\mathbf{x}_{3})>p(y_{1}=1|\mathbf{x}_{2})$ (arrow {\bf b}).
If label $1$ and $3$ are correlated,
we would like the estimations of
$p(y_{3}=1|\mathbf{x}_{2})$
and $p(y_{3}=1|\mathbf{x}_{3})$ (arrow {\bf d})
to reflect such correlations to certain extent according to how much these two labels are correlated.
This can be achieved by enforcing
$p(y_{3}=1|\mathbf{x}_{2})$
and $p(y_{3}=1|\mathbf{x}_{3})$
to satisfy similar label preference,
namely, $p(y_{3}=1|\mathbf{x}_{3})>p(y_{3}=1|\mathbf{x}_{2})$ with certain high probability
according to the correlation between two labels.
As a by-product, we have $p(y_{3}=1|\mathbf{x}_{3})>p(y_{1}=1|\mathbf{x}_{2})$ (arrow {\bf c})
and therefore enforce label preference across label and instances to follow the correlation between labels.
In summary,
we can tackle the challenge in two steps:
\begin{itemize}
\item estimate the correlations between labels accurately
\item optimize microAUC by estimating label relevance according to the label correlations estimated above.
\end{itemize}
We describe the second step first.
A representation of label correlation is needed.
Here we model all pairs of label correlation using the partial correlation matrix of labels.

\begin{defi}[Partial Correlations]
Partial correlation between labels $\ell$ and $\ell^{\prime}$
is the correlation between two labels given the other labels.
\end{defi}

The partial correlations can be captured by an $l\times l$ symmetric matrix $\Omega^{-1}$,
which is called precision matrix in multivariate statistics.
To estimate the relevance scores of the labels $Y$ 
following the estimated $\Omega^{-1}$, we set up an optimization objective that combines two goals.
The first goal is to minimize certain loss function employed in model combination algorithms.
For example, the loss function in simple averaging given in Eq.(\ref{eq:consensus_0}).
The second goal is to maximize the correlation between the label partial correlation (matrix $\Omega^{-1}$)
and the empirical label correlation (given by $Y^{\prime}Y$).
The latter goal can be formulated by the inner product of two matrices, namely, $\Tr(Y^{\prime}Y\Omega^{-1})=\Tr(Y\Omega^{-1}Y^{\prime})$.
The optimization problem is given by
\begin{eqnarray}
\displaystyle\min_{Y} & J=\displaystyle \text{\{consensus loss\}}+ \Tr(Y\Omega^{-1}Y^{\prime})
\label{eq:optimization_pccm}
\end{eqnarray}
where $Y$ is the consolidated labels.
An example of the above optimization problem is given
by taking the consensus loss as the loss function Eq.(\ref{eq:consensus_0}).
%
\begin{eqnarray}
\label{eq:consensus_1}
 \displaystyle\min_{Y}& J=\displaystyle\|\bar{Y}-Y\|^{2}+\Tr(Y\Omega^{-1}Y^{\prime})
\end{eqnarray}

Taking the derivative of $J$ with respect to
the $i$th row of $Y$, $\mathbf{y}_{i}$, we obtain
\begin{equation}
\frac{\partial J}{\partial \mathbf{y}_{i}}=-2(\mathbf{\bar y}_{i}-\mathbf{y}_{i})^{\prime}+2\Omega^{-1}\mathbf{y}_{i}^{\prime}
\end{equation}
Equating the above derivative to 0, we get
\begin{equation}
\mathbf{y}_{i}=\sum_{k=1}^{m}\mathbf{y}^{k}_{i}(\Omega^{-1}+mI_{l})^{-1}=m\mathbf{\bar y}_{i}(\Omega^{-1}+mI_{l})^{-1}
\label{eq:averaging_omega}
\end{equation}
By comparing Eq.(\ref{eq:averaging_omega}) and Eq.(\ref{eq:averaging}),
one can see that label correlation $\Omega$ is now
taken into account when producing the consolidated predictions.

Note that we assume $\Omega$ is given in the above optimization problem.
In reality, $\Omega$ is usually unknown and has to be estimated from data.
Below we show how to estimate $\Omega$ using MLE.
In order to set up an MLE problem, one needs to assume density functions for the observed data given the parameter.
Here we treat $Y=\{\mathbf{y}_{1},\dots, \mathbf{y}_{n}\}$ as the data independently generated from the normal density
\begin{equation}
\mathbf{y}_{i}\sim {\cal N}(\mathbf{0}, \Omega)=\frac{1}{C}\exp\{-\frac{1}{2}\mathbf{y}_{i}^{\prime}\Omega^{-1}\mathbf{y}_{i}\}
\end{equation}
where $C=(2\pi)^{l/2}|\Omega|^{1/2}$ is the normalization constant.
The likelihood of $Y$ given $\Omega$ is
\begin{equation}
p(Y|\Omega)=\prod_{i=1}^{n}p(\mathbf{y}_{i}|\Omega)=\frac{1}{C^{n}}\exp\{-\frac{1}{2}\sum_{i=1}^{n}\mathbf{y}_{i}^{\prime}\Omega^{-1}\mathbf{y}_{i}\}
\end{equation}
According to the MLE of the covariance matrix of multivariate Gaussian distributions,
$\Omega$ is estimated as
\begin{equation}
{\hat \Omega}_{MLE}=\frac{1}{n}Y^{\prime}Y\nonumber
\end{equation}
Now we can put the above two steps together to build the MLCM-a algorithm, as described
in Algorithm~\ref{alg:mlcma}.
\begin{algorithm}
\caption{MLCM-a}
\label{alg:mlcma}
\begin{algorithmic}[1]
\STATE {\bf Input}: Predictions from base models $\{Y^{1},\dots, Y^{m}\}$
\STATE {\bf Output}: Consolidated predictions $Y$.
\STATE Estimate $Y={\bar Y}$
\FOR{$t=1 \to T$}
\STATE Estimate covariance $\Omega=\frac{1}{n}Y^{\prime}Y$
\STATE Estimate $Y$ using Eq.(\ref{eq:averaging_omega})
\ENDFOR
\end{algorithmic}
\end{algorithm}

\section{Experiments}
\subsection{Datasets}
With 6 datasets widely used in multilabel classification community,
we demonstrate the effectiveness of the proposed methods.
Their properties are summarized in Table~\ref{tab:datasets}.
Note that these datasets have a relatively large number of labels,
it can be very time-consuming for multilabel classification models
to account for complex label correlations during training. 
\begin{table}
\caption{Datasets}
\label{tab:datasets}
\centering
\begin{tabular}{cccc}\hline
datasets & \# of instances & \# of features & \# of labels \\\hline
enron & 1702 & 1054 & 53 \\
medical & 978 & 1449 & 45\\
rcv1 subset 1& 2997 & 47337 & 101 \\
rcv1 subset 2& 2951 & 47337 & 101 \\
slashdot & 3782 & 1101 & 22 \\
bibtex & 3701 & 1995 & 159 \\\hline
\end{tabular}
\end{table}

\subsection{Evaluation Metrics}
We further include certain popular metrics to give some empirical observations
as guidance for the use of the proposed methods in practice.
For a multilabel classifier $f$, the ranking of the labels of an instance $\mathbf{x}$
is given by $\{\ell_1^{\prime},\dots, \ell_{c}^{\prime}\}$
where $f(\ell_1^{\prime},\mathbf{x})\geq f(\ell_{2}^{\prime}, \mathbf{x})\geq\dots\geq f(\ell_{c}^{\prime},\mathbf{x})$
and $f(\ell, \mathbf{x})$ is the relevance score of the label $\ell$ to $\mathbf{x}$ according to $f$.

\textbullet \hspace{.1in} {\it one error}: an error occurs when the top-ranked label is not a relevant one,
otherwise there is no error, regardless of how the other labels are ranked.
\begin{equation}
\text{\it one error}=\frac{1}{n}\sum_{i=1}^{n} \mathbbm{1}[\ell_{1}^{\prime}\not\in \mathbf{z}_{i}]
\end{equation}
where $\ell_{1}^{\prime}$ is the most relevant label to $\mathbf{x}_{i}$ according to $f$ and
$\mathbf{z}_{i}$ is the set of relevant labels of $\mathbf{x}_{i}$.
$\mathbbm{1}[\cdot]$ is 1 if and only if the statement in the brackets is true.
The lower the one error, the better an algorithm performs.

\textbullet\hspace{.1in} {\it average precision}: evaluates the precision averaged over all instances and all possible numbers of retrieved labels.
\begin{equation}
\text{\it average precision}=\frac{1}{n}\sum_{i=1}^{n}\frac{1}{c}\sum_{s=1}^{c}\frac{\{\ell_{1}^{\prime},\dots,\ell_{s}^{\prime}\}\cap \mathbf{z}_{i}}{s}
\end{equation}
where $\{\ell_{1}^{\prime},\dots,\ell_{s}^{\prime}\}$ is the top $s$ labels retrieved for instance $\mathbf{x}_{i}$
(the subscript $i$ is ignored in the retrieved labels).
The higher the average precision, the better an algorithm performs.

\subsection{Baselines}
We compare the proposed methods to two baselines.
First, 
evaluation metrics are computed for each base model,
the averaged performance of base models (denoted by BM in the sequel) are obtained as one of the baselines.
Second, we also report the performance of majority voting method (MV in the sequel).
The predictions of all base models are averaged and evaluation metrics are computed
using the averaged predictions.
By comparison of these two methods, we would be able to see how model averaging
improves the performance in the multilabel setting.
This confirm the effectiveness of ensemble method used in multilabel classification~\cite{Read09, TsoumakasKV11}.
Since we do not assume the base models have considered
label correlation in training or testing phase, while
majority voting cannot discover and exploit label correlations,
the proposed methods should be able to outperform
the base models and majority voting.

\subsection{Experiment settings}
A base model is obtained by first randomly shuffling the dataset, followed by 10-fold CV.
For each dataset, we training 10 such base models.
For each base model, one can calculate its performance using the metrics mentioned above.
The predictions of these base models are used as input to MV, MLCM-r and MLCM-a,
each of which produces consolidated predictions.
Based on the consolidated predictions, we can evaluate the performance of MV, MLCM-r and MLCM-a.
This experiment is repeated for 10 times for each dataset and the averaged performance
is reported next.

\subsection{Results}
We show the performance of the proposed algorithms and baselines in Table~\ref{tab:enron}-\ref{tab:bibtex}.
We have a couple of observations.
First,
by comparing results in the rows for BM and MV,
one can see that
combining model can boost the performance of multilabel classification,
even only using the simplest way of combination (simple averaging here).
The maximum improvements of MV over BM are 41\% and 12.8\%
for ranking loss and microAUC, respectively.
This is not surprising, as this method is widely used in ensemble multilabel classification methods
like~\cite{Read09, TsoumakasKV11, Yan2007, Schapire2000, Yu2012, Shi2011, Zhang2010}.
Second, by comparing the results of the proposed methods and simple averaging,
we observe that
simple averaging is not sufficient to fully exploit label correlations, especially when
the base models do not take the correlations into account.
The maximum improvement of either the proposed algorithms over MV is 45\% in ranking loss
and 20\% in microAUC.
Third, out of 6 tasks, MLCM-r wins MLCM-a 5 times in ranking loss, with a maximum of 12\% improvement,
and MLCM-a wins MLCM-r 4 times in microAUC, with a maximum of 5.8\% improvement.
The above comparisons show the superiority of the proposed methods over the baselines
for multilabel predictions combination tasks, and also how to choose from the proposed methods
when different metrics are considered.
Lastly, besides ranking loss and microAUC, the proposed methods also outperform the baselines with the other two metrics,
and this shows the wide applicability of the proposed methods.
\begin{table}
\centering
\footnotesize
\caption{Results on enron dataset}
\label{tab:enron}
\begin{tabular}{c|cccc}\hline\hline
\multirow{2}{*}{Methods} &\multicolumn{4}{c}{Metrics}\\\cline{2-5}
& microAUC & one error & ranking loss & avg precision \\\hline
BM & 0.7342&	0.5024&	0.2967&	0.4592\\
MV & 0.8289&	0.3398&	0.1848&	0.6020\\
MLCM-r & 0.8759&	0.6233&	{\bf 0.1003}&	0.5252\\
MLCM-a & {\bf 0.8931}&	{\bf 0.2675}&	0.1070&	{\bf 0.6556}\\\hline
\end{tabular}
\end{table}

\begin{table}
\centering
\footnotesize
\caption{Results on medical dataset}
\label{tab:yeast}
\begin{tabular}{c|cccc}\hline\hline
\multirow{2}{*}{Methods} &\multicolumn{4}{c}{Metrics}\\\cline{2-5}
& microAUC & one error & ranking loss & avg precision \\\hline
BM & 0.8887&	0.2041&	0.0989&	0.7953\\
MV & 0.9321&	0.1410& 0.0582&	0.8639\\
MLCM-r & 0.9536&	0.1327&	{\bf 0.0494}&	{\bf 0.8750}\\
MLCM-a & {\bf 0.9556}&	{\bf 0.1322}&	0.0530&	0.8649\\\hline
\end{tabular}
\end{table}

\begin{table}
\centering
\footnotesize
\caption{Results on rcv1 subset 1 dataset}
\label{tab:rcv1_1}
\begin{tabular}{c|cccc}\hline\hline
\multirow{2}{*}{Methods} &\multicolumn{4}{c}{Metrics}\\\cline{2-5}
& microAUC & one error & ranking loss & avg precision \\\hline
BM & 0.6194&	0.6036&	0.3373&	0.3218\\
MV & 0.6787&	0.4792&	0.2838&	0.4164\\
MLCM-r & 0.7867&	0.3554&	{\bf 0.2316}&	{\bf 0.5017}\\
MLCM-a & {\bf 0.8069}&	{\bf 0.3120}&	0.2605&	0.4967\\\hline
\end{tabular}
\end{table}

\begin{table}
\centering
\footnotesize
\caption{Results on rcv1 subset 2 dataset}
\label{tab:rcv1_2}
\begin{tabular}{c|cccc}\hline\hline
\multirow{2}{*}{Methods} &\multicolumn{4}{c}{Metrics}\\\cline{2-5}
& microAUC & one error & ranking loss & avg precision \\\hline
BM & 0.6220&	0.5652&	0.5652&	0.3659\\
MV & 0.6678&	0.4730&	0.4730&	0.4389\\
MLCM-r & 0.7581&	0.2955&	0.2955&	{\bf 0.5146}\\
MLCM-a & {\bf 0.8020}&	{\bf 0.2830}&	{\bf 0.2830}&	0.5073\\\hline
\end{tabular}
\end{table}

\begin{table}
\centering
\footnotesize
\caption{Results on slashdot dataset}
\label{tab:slashdot}
\begin{tabular}{c|cccc}\hline\hline
\multirow{2}{*}{Methods} &\multicolumn{4}{c}{Metrics}\\\cline{2-5}
& microAUC & one error & ranking loss & avg precision \\\hline
BM&	0.7377&	0.4875&	0.2062&	0.5856\\
MV&	0.8210&	0.4085&	0.1482&	0.6689\\
MLCM-r & {\bf 0.8782}&	0.4123&	{\bf 0.1203}&	0.6736\\
MLCM-a &0.8702&	{\bf 0.3887}&	0.1289&	{\bf 0.6800}\\\hline
\end{tabular}
\end{table}

\begin{table}
\centering
\footnotesize
\caption{Results on bibtex dataset}
\label{tab:bibtex}
\begin{tabular}{c|cccc}\hline\hline
\multirow{2}{*}{Methods} &\multicolumn{4}{c}{Metrics}\\\cline{2-5}
&  microAUC & one error & ranking loss & avg precision \\\hline
BM& 0.6620&	0.5469&	0.3095&	0.3575\\
MV& 0.7266&	0.4329&	0.2508&	0.4567\\
MLCM-r & {\bf 0.8668}&	0.4713&	{\bf 0.1599}&	0.4828\\
MLCM-a &0.8645&	{\bf 0.3790}&	0.1755&	{\bf 0.4937}\\\hline
\end{tabular}
\end{table}

\section{Related Work}
To the best of our knowledge, this work is the first attempt
to address the challenge of combining multilabel predictions of an ensemble of base models.
The proposed algorithms is different from but related to ensemble learning and  multilabel classification,
We briefly discuss these areas and how they related to this work below.
In multilabel classification, an instance have more than one label, contrasting to binary/multiclass classification
where there is only one label.
A multilabel classifier predicts the value of all labels as output.
Depending on how label relationships are dealt with, multilabel classification methods can be roughly categorized
as following.
(1) Binary Relevance.
Labels are treated as independent and prediction of each label is handled by individual binary/multiclass model.
Using this principle,
in Section~\ref{sec:bgcm} we pointed out
a naive way to combine multilabel predictions of base models.
That is to apply any prediction combination method to each label and then output the predictions on all labels.
The binary relevance paradigm does not consider label dependency and thus might be inferior to methods that consider label dependency in
terms of prediction performance.
(2) Pairwise relationship.
This category of methods model the relationships between two labels.
In~\cite{Zhang10}, they propose a method to learn label relationships using Bayesian network, which
is later utilized to learn a binary classifiers for each label given that label's parent labels.
(3) Powerset Methods.
This set of methods try to fully consider all possible co-occurrence of labels.
In particular, a set of labels is considered as a class, and the multilabel problem
is reduced to a multiclass problem.
A classifier needs to map an instance into a class, which is a set of labels.
The drawback of these methods is that the number of label sets increases exponentially in the number of labels.
Example algorithms in this category include those in~\cite{Read09, TsoumakasKV11}.

There have been an extensive study of ensemble methods,
which combines the knowledge of multiple models to improve performance.
\cite{zhou12} provides an excellent review of ensemble methods, here we discuss those methods that are
only relevant to this paper.
The simplest ensemble method is majority voting.
In~\cite{Breiman1996}, bootstrap sampling is used to create multiple copies of training data
to derive an ensemble of models.
It is shown that bagging improves performance via reduction in variance.
Another famous ensemble method is boosting~\cite{Schapire2002}, which
builds the ensemble via sequential training of base models
to exploit model correlation.
The success of boosting can be explained by the margin theory~\cite{Schapire97, Reyzin06}.
Ensemble methods have some important applications, such as in
classification with skew class distribution~\cite{kun06}
data stream mining\cite{jing07}, knowledge transfer~\cite{jing08}

Combining predictions without access to training or test data has been researched for at least a decade.
\cite{strehl03} is probably the most well known paper in this topic.
They present three methods, CSPA, HGPA and MCLA for cluster ensemble.
In~\cite{wang09} they propose a Bayesian framework to infer the ground truths of the instances
given the predictions of base models.
In~\cite{li07}, matrix factorization is employed to obtain a low dimensional representation
of the instances given the similarity matrix derived from the predictions.
In~\cite{jing08}, they propose BGCM, which maximizes the consensus among models.
None of these methods can directly address to the multilabel prediction combination problem.

In~\cite{Zhang2011},
they explore the idea of modeling the diversity of multiple kernels and the correlations of pairwise labels.
This idea is similar to the proposed methods
and thus can not be applied to the problem we are solving in this paper.
In~\cite{Tang09}, they treat the learning of a model for a label as a stand-along task.
Then their algorithm learns a linear combination of multiple kernels for each task.
The only thing that related multilabel is the set of kernels to be combined for each label.
The drawback of these methods and those proposed in \cite{huang12, Zhang2010, Yu2012, Shi2011}
assume that training and test data are available and therefore cannot address the challenge of this paper.

\section{Conclusion}
In this paper, we aim at combining multilabel predictions from multiple models.
The challenge is how to exploit label correlations to optimize a certain performance metric when consolidating predictions.
Existing multilabel ensemble algorithms fail to do so.
We address the challenge via two methods: MLCM-r and MLCM-a.
The former uses random walk in the label space to explicitly infer label correlation,
which in turn results in consolidated multilabel predictions optimized for ranking loss.
The latter uses an optimization framework to estimate the partial label correlations, which
regularizes predictions consolidation to optimize microAUC.
We analyze both algorithms to establish these optimal properties.
Experimental results affirmatively demonstrate the superiority of the proposed algorithms.
\bibliography{paper}
\bibliographystyle{plain}
\end{document}